\documentclass[letterpaper, 10 pt, journal, twoside]{IEEEtran}  

\IEEEoverridecommandlockouts                              

\usepackage{amsmath} 
\usepackage{amssymb}  
\usepackage{bm}
\usepackage{multirow}
\usepackage{makecell}
\usepackage{bbding}

\usepackage{graphics} 
\usepackage{epsfig} 

\usepackage{mathtools}
\usepackage{enumitem}
\usepackage{times} 

\usepackage[utf8]{inputenc} 
\usepackage[T1]{fontenc}    
\usepackage{hyperref}       
            
\usepackage{url}            
\usepackage{booktabs}       
\usepackage{amsfonts}       
\usepackage{nicefrac}       
\usepackage{microtype}      
\usepackage{xcolor}         
\usepackage{cleveref}
\usepackage{hyperref}
\usepackage{graphicx}
\usepackage{float}
\usepackage{multirow}
\usepackage{diagbox}

\usepackage{algorithm}
\usepackage{algpseudocode}
\usepackage{arydshln}
\usepackage{marvosym}

\usepackage{multirow}
\usepackage{listings}
\usepackage{graphicx}
\usepackage{siunitx} 
\usepackage{pifont}
\usepackage{adjustbox}

\usepackage[caption=false,font=normalsize,labelfont=sf,textfont=sf]{subfig}
\definecolor{dkgreen}{rgb}{0,0.6,0}
\definecolor{gray}{rgb}{0.5,0.5,0.5}
\definecolor{mauve}{rgb}{0.58,0,0.82}
\begin{document}

\markboth{IEEE Robotics and Automation Letters. Preprint Version. ACCEPTED September, 2025}{Li \MakeLowercase{\textit{et al.}}: Enhancing Indoor Occupancy Prediction via Sparse Query-Based Multi-Level Consistent Knowledge Distillation} 

\author{Xiang Li,
Yupeng Zheng$^*$,
Pengfei Li,
Yilun Chen,
Ya-Qin Zhang,
Wenchao Ding$^\dagger$
\thanks{Manuscript received: June 26, 2025; Revised: September 8, 2025; Accepted: September 21, 2025.}
\thanks{This paper was recommended for publication by Editor Pascal Vasseur upon evaluation of the Associate Editor and Reviewers’ comments.}
\thanks{This work was supported by the Tsinghua-TARS Special Program for the Deep Collaboration in Embodied Intelligence. \textit{($^*$Project leader: Yupeng Zheng. $^\dagger$Corresponding author: Wenchao Ding.)}} 
\thanks{Xiang Li is with the College of AI, Tsinghua University, Beijing 100083, China and TARS, Shanghai 200233, China.}
\thanks{Yupeng Zheng is with the Institute of Automation, Chinese Academy of Sciences, Beijing 100190, China and TARS, Shanghai 200233, China.}
\thanks{Pengfei Li is with the AIR, Tsinghua University, Beijing 100084, China and TARS, Shanghai 200233, China.}
\thanks{Yilun Chen is with the TARS, Shanghai 200233, China.}
\thanks{Ya-Qin Zhang is with the AIR, Tsinghua University, Beijing 100084, China.}
\thanks{Wenchao Ding is with the College of Intelligent Robotics and Advanced Manufacturing, Fudan University, Shanghai 200001, China and TARS, Shanghai 200233, China (e-mail: dingwenchao@fudan.edu.cn).
}
\thanks{Digital Object Identifier (DOI): see top of this page.}
}

\title{
Enhancing Indoor Occupancy Prediction via\\Sparse Query-Based Multi-Level Consistent Knowledge Distillation
}

\maketitle


\begin{abstract}

Occupancy prediction provides critical geometric and semantic understanding for robotics but faces efficiency-accuracy trade-offs. 
Current dense methods suffer computational waste on empty voxels, while sparse query-based approaches lack robustness in diverse and complex indoor scenes. 
In this paper, we propose DiScene, a novel sparse query-based framework that leverages multi-level distillation to achieve efficient and robust occupancy prediction.
In particular, our method incorporates two key innovations: (1) a Multi-level Consistent Knowledge Distillation strategy, which transfers hierarchical representations from large teacher models to lightweight students through coordinated alignment across four levels, including encoder-level feature alignment, query-level feature matching, prior-level spatial guidance, and anchor-level high-confidence knowledge transfer and (2) a Teacher-Guided Initialization policy, employing optimized parameter warm-up to accelerate model convergence.
Validated on the Occ-Scannet benchmark, DiScene achieves 23.2 FPS without depth priors while outperforming our baseline method, OPUS, by 36.1\% and even better than the depth-enhanced version, OPUS$\dagger$. 
With depth integration, DiScene$\dagger$ attains new SOTA performance, surpassing EmbodiedOcc by 3.7\% with 1.62$\times$ faster inference speed.
Furthermore, experiments on the Occ3D-nuScenes benchmark and in-the-wild scenarios demonstrate the versatility of our approach in various environments.
Code and models can be accessed at \url{https://github.com/getterupper/DiScene}.

\end{abstract}



\section{Introduction}
\IEEEPARstart{O}{ccupancy} prediction has gained significant attention in robotics society due to its ability to provide fine-grained geometric and semantic information~\cite{popovic2021volumetric, wang2021learning}. Its objective is to estimate the occupancy status of each voxel and their semantic labels within an entire scene from limited observations. Current mainstream methods typically employ explicit 3D spatial modeling (\textit{e.g.}, dense voxels~\cite{cao2022monoscene, li2023voxformer}, Bird's-Eye View~\cite{huang2021bevdet, li2023bevdepth}, Tri-Perspective View~\cite{huang2023tri}), where most computational resources are consumed by empty voxel calculations, resulting in inefficiency. 
Alternative sparse query-based approaches~\cite{wang2024opus} simultaneously perform spatial occupancy regression and semantic label classification, thereby significantly accelerating inference speeds.

\begin{figure}[t]
  \centering
  \includegraphics[width=0.95\linewidth]{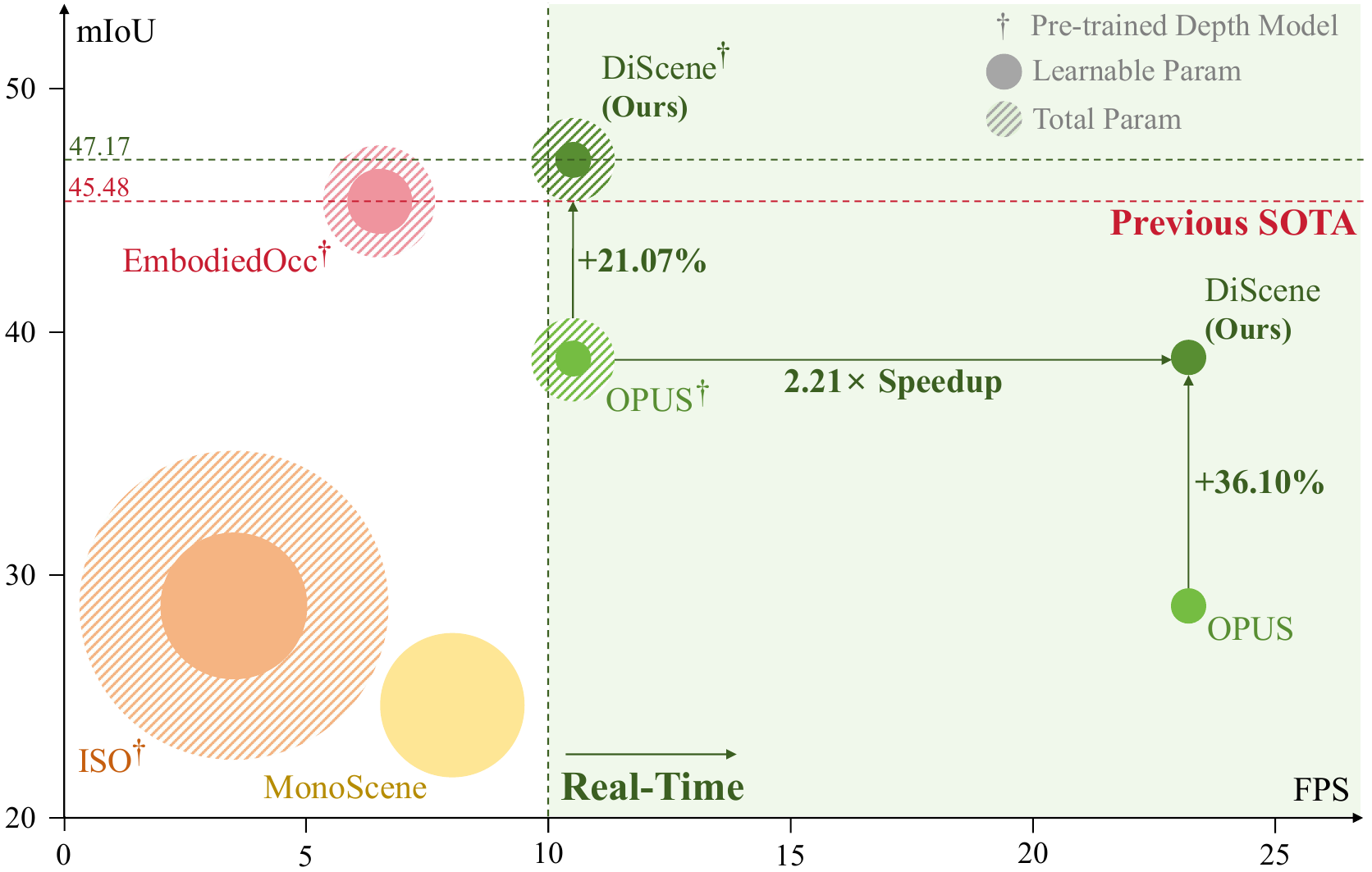}
  \vspace{-3mm}
  \caption{We compare DiScene with existing indoor occupancy prediction methods in terms of speed and accuracy. All models are evaluated on the Occ-ScanNet~\cite{yu2024monocular} validation set and inference speeds are measured on one NVIDIA A800 GPU w/o TensorRT. The size of the circle represents the model’s size.}
  \label{fig:teaser}
  \vspace{-6mm}
\end{figure}

However, these sparse methods underperform in diverse and complex indoor scenes due to insufficient geometric information. 
Introducing additional vision foundation models~\cite{yang2024depthanythingv2, hu2024metric3dv2} to mitigate the ambiguity, in turn, increases latency and compromises real-time performance. Hence, current indoor occupancy prediction methods still fail to achieve a satisfactory balance between model performance and inference speed.

To address this critical limitation, we propose \textbf{DiScene}, a novel distillation framework specifically designed for sparse query-based occupancy prediction. We argue that there are several challenges that prevent traditional distillation methods from achieving optimal gains: large feature discrepancy between teacher and student models impedes effective knowledge transfer; there is no natural one-to-one correspondence between teacher and student predictions for vanilla logit- or feature-based distillation; directly distilling both spatial distributions and feature representations from sparse teacher queries introduces excessive learning complexity. To overcome the above challenges, we pioneer a hierarchical distillation strategy that establishes coordinated knowledge transfer between teacher and student models and progressively incorporates guidance information, thus effectively reducing the difficulty of the distillation learning process while maximizing the efficacy of knowledge transfer.

Specifically, we adopt the sparse-centric model OPUS~\cite{wang2024opus} as our baseline and introduce a \textbf{Multi-level Consistent Knowledge Distillation} strategy, comprising  
(1) \textit{Encoder-level Feature Alignment:} We address feature discrepancy caused by heterogeneous encoders through encoder-level alignment loss, enabling effective distillation between teacher and student models.
(2) \textit{Query-level Distillation:} We utilize the Hungarian algorithm to establish optimal bipartite matching between student and teacher predictions, allowing coarse feature-based knowledge transfer.
(3) \textit{Prior-level Distillation:} We provide the teacher query positions as spatial priors to the student. This addresses unstable bipartite matching resulting from divergent spatial distributions of query embeddings, facilitating more focused feature representation learning.
(4) \textit{Anchor-level Distillation:} We sample anchor points from ground-truth occupied voxels and provide them to both models. This ensures selective transfer of high-confidence knowledge, thereby preventing knowledge contamination from low-quality teacher predictions and further enhancing feature representation learning. Moreover, we propose a \textbf{Teacher-Guided Initialization} policy, which utilizes well-optimized teacher parameters to accelerate model convergence as free lunch.

DiScene achieves SOTA occupancy prediction performance and real-time inference on the challenging Occ-Scannet benchmark~\cite{yu2024monocular}. 
As demonstrated in Fig.~\ref{fig:teaser}, relying solely on distillation and initialization strategy, DiScene maintains a 23.2 FPS inference speed while outperforming the baseline method OPUS by 36.1\% and delivers comparable performance to OPUS$\dagger$, which leverages pre-trained depth models. 
When incorporating depth priors, our enhanced DiScene$\dagger$ surpasses the previous SOTA method EmbodiedOcc~\cite{wu2024embodiedocc} by 3.7\% while sustaining inference speeds above 10 FPS.

Moreover, on the Occ3D-nuScenes~\cite{tian2023occ3d} benchmark, our strategy improves performance by 6.9\%, exhibiting robust performance across both indoor and outdoor robotic perception scenarios. We further validate the generality and versatility of our approach on self-collected in-the-wild datasets.

Our main contributions are as follows: 
\begin{itemize}
    \item We propose DiScene, a sparse query-based distillation framework that bridges the accuracy-efficiency gap prevalent in existing indoor occupancy prediction methods.
    \item We propose a Multi-level Consistent Knowledge Distillation strategy that ensures effective knowledge transfer across multiple complementary levels.
    \item We introduce a Teacher-Guided Initialization policy that accelerates model convergence at no additional costs.
    \item We demonstrate the effectiveness and robustness of our method through extensive experiments across indoor and outdoor benchmarks, with additional validation on in-the-wild scenarios.
\end{itemize}

\section{Related Work}

\subsection{Occupancy Prediction}

Occupancy prediction has achieved notable progress in recent years. Conventional 3D~\cite{cao2022monoscene, wei2023surroundocc, tian2023occ3d, tong2023scene, li2023fb, pan2024renderocc} or 4D~\cite{li2025semi, gu2024dome, jin2025occtens} methods predominantly employ dense voxels as feature representation~; however, such an approach incurs heavy and redundant computational costs. 
Consequently, recent research in outdoor driving scenarios has seen the emergence of numerous acceleration techniques utilizing alternative representations, such as Bird's-Eye View~\cite{yu2023flashocc, hou2024fastocc}, Tri-Perspective View~\cite{huang2023tri}, 3D Gaussians~\cite{huang2024gaussianformerv1, huang2025gaussianformerv2} and sparse 3D queries~\cite{li2023voxformer, liu2024fully, wang2024opus}.

On the contrary, similar efforts have not yet been observed in indoor scenarios. Methods like ISO~\cite{yu2024monocular} and EmbodiedOcc~\cite{wu2024embodiedocc} leverage pre-trained depth models~\cite{yang2024depthanything, yang2024depthanythingv2} to estimate depth information, which is then fused with scene features to enhance model performance by mitigating depth ambiguity. Nevertheless, the incorporation of such depth models substantially increases inference overhead, hindering their practical deployment in the real world.

In this letter, we attempt to strike a balance between performance and real-time inference for indoor occupancy prediction. Our solution adopts a sparse query-based architecture as the primary framework while integrating knowledge distillation to boost performance without introducing additional costs.

\subsection{Knowledge Distillation}

As a classical method for model compression and accuracy enhancing, the concept of knowledge distillation was first introduced by~\cite{hinton2015distilling}, where students are trained to mimic the soft label predictions of teachers. According to the objective of mimicking, subsequent works can be broadly categorized into two types, distilling from output logits~\cite{zhang2018deep, mirzadeh2020improved, zhao2022decoupled, yang2024learning, yang2025adaptive} and intermediate features~\cite{heo2019comprehensive, tian2019contrastive, zhang2020task, pham2024frequency, fan2024scalekd}. Researchers have applied knowledge distillation to various vision tasks and modality and lead to consistent effectiveness, including image generation~\cite{jin2021teachers, zhang2022wavelet}, 2D semantic segmentation~\cite{liu2019structured}, 2D object detection~\cite{guo2021distilling, li2022knowledge, wang2024kd}, LiDAR semantic segmentation~\cite{hou2022point} and 3D object detection~\cite{chong2022monodistill, zhang2023pointdistiller}.

Prior works such as SCPNet~\cite{xia2023scpnet} and MonoOcc~\cite{zheng2024monoocc} have adopted this strategy for occupancy prediction, transferring geometric and semantic knowledge from multi-frame teachers to single-frame students. However, these methods employ dense 3D feature representation, which makes it easier for student models to imitate teachers due to the explicit correspondence between voxels. The application of knowledge distillation to sparse queries for occupancy prediction remains an unexplored and challenging task.
\begin{figure*}
  \centering
  \includegraphics[width=0.9\textwidth]{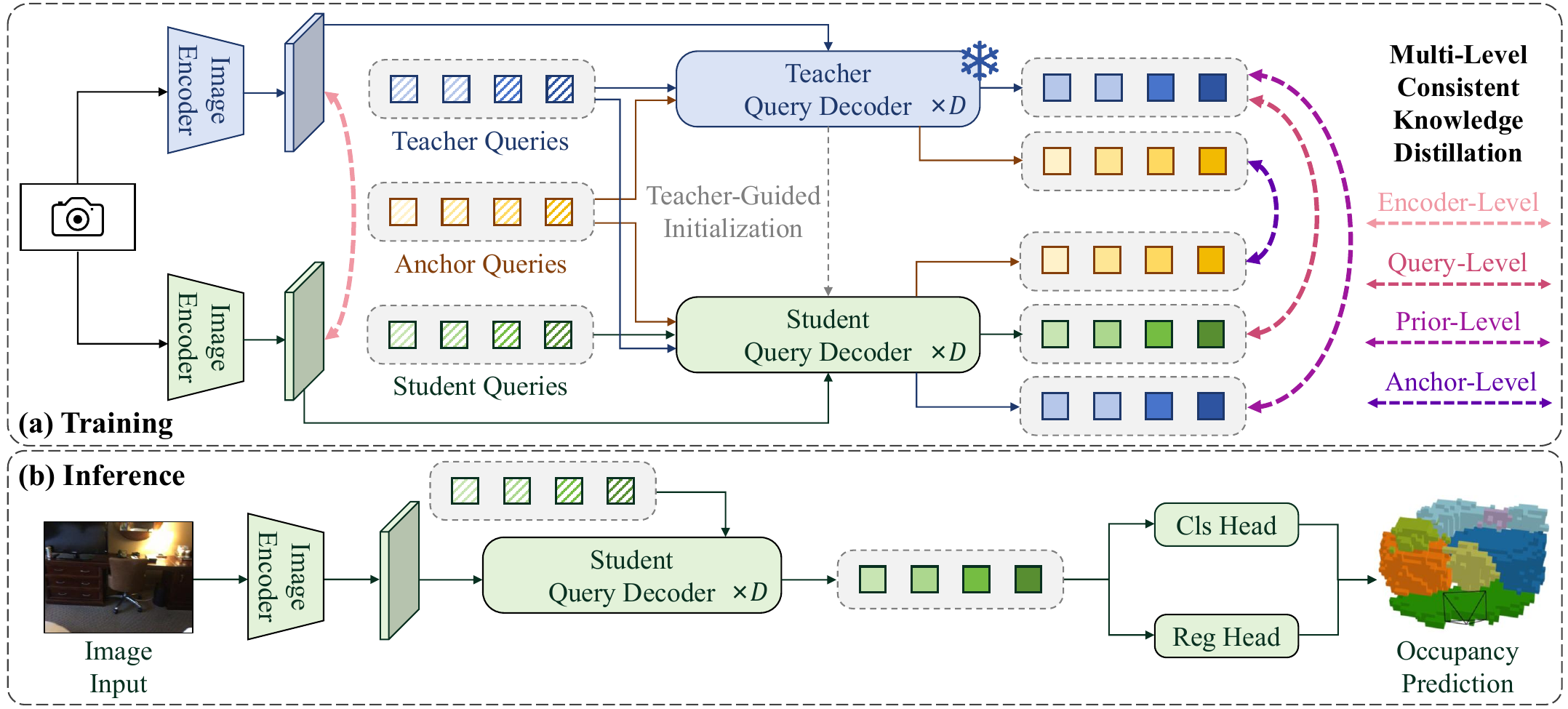}
  \vspace{-3mm}
  \caption{(a) The illustration of our proposed knowledge distillation strategies. (b) The architecture of our primary framework. Best viewed in color.}
  \vspace{-6mm}
  \label{fig:pipeline}
\end{figure*}

\section{Methodology}

\subsection{Preliminaries}
\label{subsec: model arch}

\textbf{Problem Formulation.} Following OPUS~\cite{wang2024opus}, we reformulate occupancy prediction as a set-to-set matching task to better leverage the sparsity inherent in indoor scenes. Given $M$ occupied voxels of the current scene, we denote them as a ground-truth set $\mathcal{S}^{\text{GT}}=\{\mathcal{P}^{\text{GT}}, \mathcal{C}^{\text{GT}}\}=\{p_i^{\text{GT}}, c_i^{\text{GT}}\}_{i=1}^{M}$, where $p_i^{\text{GT}}$ denotes the 3D coordinates of a voxel center, and $c_i^{\text{GT}}$ represents its corresponding semantic class. Our model predicts $M'$ point positions and semantic classes, denoted as the prediction set $\mathcal{S}^{\text{Pred}}=\{\mathcal{P}^{\text{Pred}}, \mathcal{C}^{\text{Pred}}\}$, where $M$ and $M'$ are not necessarily equal. Consequently, the goal of our method is to minimize the discrepancy between the distributions of the two sets, $\mathcal{S}^{\text{GT}}$ and $\mathcal{S}^{\text{Pred}}$.

\textbf{Vanilla Pipeline.} As illustrated in Fig.~\ref{fig:pipeline} (b), our baseline method leverages a transformer encoder-decoder architecture. Taking a set of $N$ learnable queries $\{q_i\}_{i=1}^{N}$ as input, the model utilizes an image encoder to extract 2D features from the image input, and subsequently employs a multi-layer decoder to iteratively refine the queries using image features. At the end of each decoder layer, a regression and a classification head are used to output updated position and semantic predictions. To address the computational costs incurred by an excessive number of queries, each query $q_i$ predicts the distribution of a point set $\{r_{i,j}\}_{j=1}^{R}$, where $R$ denotes the number of points generated from one query, which is progressively increased across successive decoder layers in a coarse-to-fine manner.

\textbf{Depth Branch.} Building upon this, we introduce a simple yet effective depth branch inspired by~\cite{wu2024embodiedocc}, leveraging depth predictions from a pre-trained depth model as prior information. For a given query $q$ and its corresponding point set center $\bar{r}$, We project $\bar{r}$ into the camera space and subsequently onto the image plane, yielding its projected depth value $d_{q}$ and corresponding 2D image coordinates $(u, v)$. Letting $\tilde{I}_{d}$ denote the 2D depth prediction from the pre-trained depth model, we thus obtain the prior depth value at this location as $d_{p}=\tilde{I}_{d}(u, v)$. We then encode the projected and prior depth values into a depth-prior feature $f_{d}$ using a simple MLP, which is eventually used to enhance the query feature $f$ via channel-wise addition:
\begin{equation}
f_d=\text{MLP}(d_q,d_p),\ \hat{f}=f+f_d.
\end{equation}

\textbf{Loss Function.} Both the vanilla and depth-prior settings utilize identical loss functions. We employ the Chamfer distance loss~\cite{fan2017point} to supervise the position predictions $\mathcal{P}^{\text{Pred}}$. For semantic supervision, a matched set of ground-truth semantic labels $\hat{\mathcal{C}}^{\text{GT}}$ is assigned to the predicted semantic labels $\mathcal{C}^{\text{Pred}}$ via nearest-neighbor matching. The semantic predictions are then optimized using focal loss~\cite{lin2017focal}. Therefore, the overall loss function for the task can be formulated as:
\begin{equation}
\mathcal{L}_{task}=\sum_{d=1}^{D}\mathcal{L}_{CD}(\mathcal{P}^{\text{Pred}}_{d}, \mathcal{P}^{\text{GT}})+\mathcal{L}_{focal}(\mathcal{C}^{\text{Pred}}_{d},\hat{\mathcal{C}}^{\text{GT}}_{d}),
\end{equation}
where $D$ is the number of decoder layers.

\subsection{Multi-Level Consistent Knowledge Distillation}
\label{subsec: MCKD}

\subsubsection{Encoder-Level Feature Alignment}
\label{subsec: EFA}

In practice, our student and teacher models employ heterogeneous image encoders with different scales, leading significant divergence in their image feature representations. Since query features are substantially influenced by encoder outputs, we empirically find that this discrepancy largely hinders effective knowledge transfer and even compromises student performance. To cope with this issue, we adopt a simple yet effective feature alignment loss:

\begin{equation}
\mathcal{L}_{EFA}=\mathcal{L}_{\text{MSE}}(\frac{F^S}{||F^S||_2}, \frac{F^T}{||F^T||_2}),
\end{equation}
where $F^S$ and $F^T$ denotes student and teacher image features.

\subsubsection{Query-Level Distillation}
\label{subsec: QL}

A straightforward approach for knowledge distillation is to directly align the predictions of corresponding queries between the two models. However, our student and teacher queries lack ordered one-to-one correspondence, presenting a fundamental challenge for direct application. To resolve this misalignment, we establish an optimal bipartite matching $\hat{\sigma}$ between the $N$ student queries $\{q^S_i\}_{i=1}^{N}$ and teacher queries $\{q^T_i\}_{i=1}^{N}$ using the Hungarian algorithm~\cite{kuhn1955hungarian}. In practice, we employ the L2 distance between the point set centers from the student and teacher queries as the pair-wise matching cost of the cost matrix:
\begin{equation}
c_{ij}=c(q_i^S, q_j^T) = ||\bar{r}^S_i-\bar{r}_j^T||_2.
\end{equation}
This matching ensures consistent pairing of teacher-student predictions. Thus, the query-level distillation loss can be formulated as:
\begin{equation}
\mathcal{L}_{QL}=\frac{1}{N}\sum_{d=1}^{D}\sum_{i=1}^{N}\mathcal{L}_{match}(q^S_i, q^T_{\hat{\sigma}_i}).
\end{equation}

Since each query governs multiple point positions and semantics, we investigate two granularities of $\mathcal{L}_{match}$, respectively termed fine-grained logit-based distillation and coarse feature-based distillation, denoted as $\mathcal{L}_{FLD}$ and $\mathcal{L}_{CFD}$. For $\mathcal{L}_{FLD}$, we supervise the 3D coordinates $r_{i,j}$ and output semantic logits $c_{i,j}$ for all $R$ points within the point set associated with a matched query pair $(q^S_i, q^T_{\hat{\sigma}_i})$. While for $\mathcal{L}_{CFD}$, we only supervise the point set center position $\bar{r}_i$ and the query feature $f_i$. We use L1 loss, Kullback-Leibler Divergence loss and MSE loss for position, semantic and feature distillation, respectively:
\begin{equation}
\begin{aligned}
\mathcal{L}_{FLD}=\frac{1}{R}\sum_{j=1}^{R}\mathcal{L}_{\text{L1}}(r_{i,j}^S, r^T_{\hat{\sigma}_i,j}) + \mathcal{L}_{\text{KL}}(c_{i,j}^S, c_{\hat{\sigma}_i,j}^T), \\
\mathcal{L}_{CFD}=\mathcal{L}_{\text{L1}}(\bar{r}_i^S, \bar{r}_{\hat{\sigma}_i}^T) + \mathcal{L}_{\text{MSE}}(\frac{f_i^S}{||f_i^S||_2}, \frac{f_{\hat{\sigma}_i}^T}{||f_{\hat{\sigma}_i}^T||_2}).
\end{aligned}
\end{equation}

Empirically, we observe that coarse feature-based distillation facilitates more effective knowledge transfer to the student model. Therefore, our final query-level distillation loss can be represented as follows:
\begin{equation}
\begin{aligned}
\mathcal{L}_{QL}\!&=\!\frac{1}{N}\sum_{d=1}^{D}\sum_{i=1}^{N}\mathcal{L}_{CFD}(q^S_i, q^T_{\hat{\sigma}_i})\\
\!&=\!\frac{1}{N}\sum_{d=1}^{D}\sum_{i=1}^{N}\mathcal{L}_{\text{L1}}(\bar{r}_i^S,\!\bar{r}_{\hat{\sigma}_i}^T)\!+\!\mathcal{L}_{\text{MSE}}(\frac{f_i^S}{||f_i^S||_2},\!\frac{f_{\hat{\sigma}_i}^T}{||f_{\hat{\sigma}_i}^T||_2}).
\end{aligned}
\end{equation}

\begin{table*}[ht]
\scriptsize
\caption{Quantitative comparison on the Occ-ScanNet validation set}
\vspace{-3mm}
\centering
\begin{tabular}{l|c|cc|ccccccccccc|c} 
\toprule
Method & PDM & IoU & mIoU 
       & \rotatebox{90}{ceiling}
       & \rotatebox{90}{floor}
       & \rotatebox{90}{wall}
       & \rotatebox{90}{window}
       & \rotatebox{90}{chair}
       & \rotatebox{90}{bed}
       & \rotatebox{90}{sofa}
       & \rotatebox{90}{table}
       & \rotatebox{90}{tvs}
       & \rotatebox{90}{furniture}
       & \rotatebox{90}{objects}
       & FPS \\
\hline
MonoScene~\cite{cao2022monoscene} & $\times$ & 41.60 & 24.62 & 15.17 & 44.71 & 22.41 & 12.55 & 26.11 & 27.03 & 35.91 & 28.32 & 6.57 & 32.16 & 19.84 & 8.0 \\
ISO$\dagger$~\cite{yu2024monocular} & DAv1 & 42.16 & 28.71 & 19.88 & 41.88 & 22.37 & 16.98 & 29.09 & 42.43 & 42.00 & 29.60 & 10.62 & 36.36 & 24.61 & 3.5 \\
EmbodiedOcc$\dagger$~\cite{wu2024embodiedocc} & DAv2 & \textbf{53.95} & \underline{45.48} & \underline{40.90} & \textbf{50.80} & \textbf{41.90} & \underline{33.00} & \underline{41.20} & \underline{55.20} & \underline{61.90} & \underline{43.80} & \underline{35.40} & \textbf{53.50} & \textbf{42.90} & 6.5 \\
OPUS~\cite{wang2024opus} & $\times$ & 35.58 & 28.70 & 15.37 & 37.75 & 20.60 & 18.64 & 26.43 & 44.55 & 45.63 & 30.79 & 14.63 & 35.80 & 25.49 & \textbf{23.2} \\
OPUS$\dagger$~\cite{wang2024opus} & DAv2 & 45.62 & 38.96 & 39.06 & 45.04 & 34.97 & 28.63 & 35.92 & 49.27 & 54.39 & 37.93 & 23.93 & 45.04 & 34.42 & \underline{10.5} \\
\hline
DiScene (Ours) & $\times$ & 43.68 & 39.06 & 29.66 & 45.28 & 28.70 & 28.73 & 35.90 & 53.13 & 56.89 & 39.90 & 30.07 & 44.97 & 36.38 & \textbf{23.2} \\
DiScene$\dagger$ (Ours) & DAv2 & \underline{51.99} & \textbf{47.17} & \textbf{45.21} & \underline{50.63} & \underline{40.38} & \textbf{36.73} & \textbf{42.28} & \textbf{59.68} & \textbf{62.04} & \textbf{45.60} & \textbf{41.17} & \underline{52.42} & \underline{42.72} & \underline{10.5} \\
\bottomrule
\multicolumn{16}{l}{$\dagger$ represents the result with pre-trained depth model, denoted as PDM. DAv1 and DAv2 are short for Depth Anything v1~\cite{yang2024depthanything} and v2~\cite{yang2024depthanythingv2} respectively.}\\
\end{tabular}
\label{tab: quantitative comparison}
\vspace{-6mm}
\end{table*}

\subsubsection{Prior-Level Distillation}
\label{subsec: PL}

Given that the student query embeddings are randomly initialized, their spatial distribution inherently diverges from the well-optimized teacher query embeddings. This discrepancy can cause unstable and suboptimal bipartite matching during early training phases, which diminishes the effectiveness of query-level distillation and impedes model convergence. Since our coarse feature-based distillation aims to transfer the spatial distributions and feature representations of teacher queries, we decouple this process by first aligning spatial distributions between the two models using spatial priors as guidance, thus enabling the student to focus on feature learning and alleviating the mismatch problem. Based on this insight, we propose prior-level distillation.

Specifically, we input the teacher query embeddings into the student model to obtain an additional group of prior queries $\{q^P_i\}_{i=1}^{N}$ and their predictions. Since $\{q^P_i\}_{i=1}^{N}$ and $\{q^T_i\}_{i=1}^{N}$ share identical initialization distributions, we approximate their consistency and establish pairwise correspondences between the two sets of queries. This approach omits the bipartite matching process and further reduces the training time. Therefore, our prior-level distillation loss can be represented as:
\begin{equation}
\mathcal{L}_{PL}=\frac{1}{N}\sum_{d=1}^{D}\sum_{i=1}^{N}\mathcal{L}_{CFD}(q^P_i, q^T_{i}).
\end{equation}

\subsubsection{Anchor-Level Distillation}
\label{subsec: AL}

While our previous strategy enhances feature representation learning, directly distilling low-confidence predictions of the teacher may be harmful to the student model. Rather than manually filtering these suboptimal outputs, we introduce anchor-level distillation to ensure high-quality knowledge transfer. To achieve this goal, we sample $N$ anchor points from the ground-truth set $\{\mathcal{P}^{\text{GT}}, \mathcal{C}^{\text{GT}}\}$ with rebalanced weight according to the frequency distribution across different semantic classes. These anchors initialize the spatial distribution of a set of anchor queries, which are then fed into both models, obtaining updated student anchor queries $\{a_i^S\}_{i=1}^{N}$ and teacher anchor queries $\{a_i^T\}_{i=1}^{N}$.

This approach simultaneously guarantees spatial distribution consistency and restricts distillation exclusively to the high-confidence predictions of the teacher at anchor locations, thus establishing robust knowledge transfer by distilling only the most reliable knowledge representations. Analogous to our prior-level distillation strategy, knowledge transfer between corresponding anchor queries bypasses the need for bipartite matching due to their shared initialization, written as:

\begin{equation}
\mathcal{L}_{AL}=\frac{1}{N}\sum_{d=1}^{D}\sum_{i=1}^{N}\mathcal{L}_{CFD}(a^S_i, a^T_{i}).
\end{equation}

\subsubsection{Distillation Loss}
\label{subsec: loss}
To sum up, the overall loss function for knowledge distillation can be formulated as:
\begin{equation}
\mathcal{L}_{distill}\!=\!\lambda_1\mathcal{L}_{EFA}+\lambda_2\mathcal{L}_{QL}+\lambda_3\mathcal{L}_{PL}+\lambda_4\mathcal{L}_{AL}.
\end{equation}

\subsection{Teacher-Guided Initialization}
\label{subsec: TGI}

Inspired by~\cite{li2024distilling}, we empirically find that the parameters of the decoder layers in the teacher model also serve as a source of knowledge. Despite employing heterogeneous encoders, the spatial and feature representations within the decoders exhibit inherent cross-model consistency. By initializing the student decoder with pre-trained weights from the teacher decoder, we significantly accelerate convergence while obtaining performance gains at no additional computational cost.
\section{Experiment}

\subsection{Experimental Setup}
\subsubsection{Benchmark}
We adopt Occ-ScanNet~\cite{yu2024monocular} as the indoor occupancy prediction benchmark, which provides voxelized scenes in $60\times60\times36$ grids with $0.08\text{m}$ resolution, representing $4.8\text{m}\times4.8\text{m}\times2.88\text{m}$. 
Each voxel is annotated with 12 classes (11 semantic classes and 1 empty). 
Following common practices, we use mIoU and IoU as evaluation metrics.

\subsubsection{Implement Details} 
\label{sec: implement}
For teacher model, we employ InternImage-XL~\cite{wang2023internimage} as the image backbone. The input image is resized to a resolution of $480 \times 640$. 
For student model, we adopt a lightweight encoder ResNet-50~\cite{he2016deep}.
We train the model for 10 epochs on 8 A800 GPUs with a total batch size of 8 using the AdamW~\cite{loshchilov2019decoupled} optimizer. We set the learning rate to $2\times10^{-4}$ and the hyperparameters as follows: $\lambda_1=1, \lambda_2=0.2, \lambda_3=0.2, \lambda_4=0.5$.

\subsection{Quantitative and Qualitative Results}

\begin{figure}[ht]
  \centering
  \includegraphics[width=0.95\linewidth]{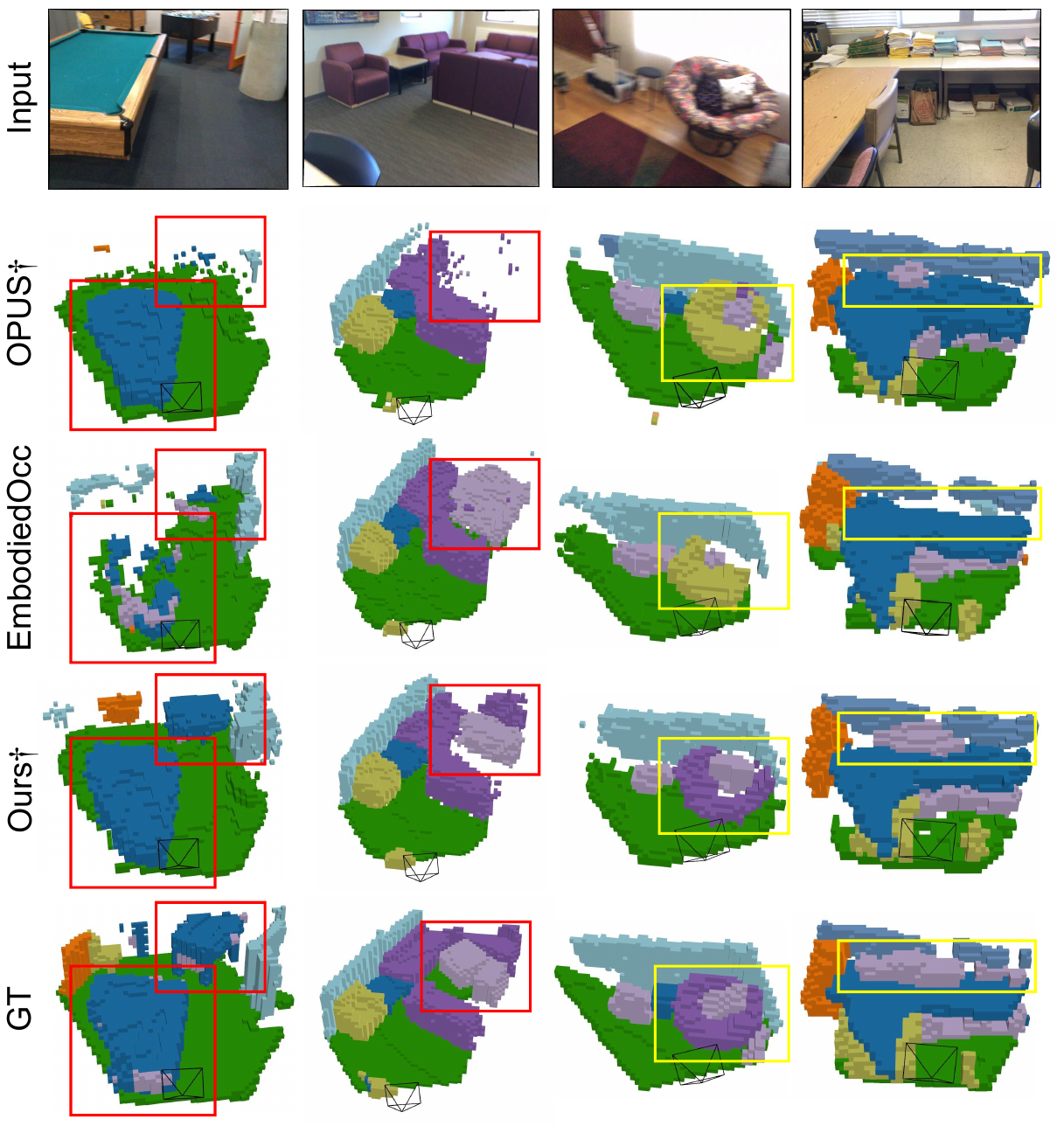}
  \vspace{-3mm}
  \caption{Qualitative results of occupancy prediction on the Occ-ScanNet validation set. Compared with existing methods, DiScene demonstrates superior geometric awareness and semantic comprehension, visually highlighted by red and yellow boxes respectively.}
  \label{fig:vis}
  \vspace{-8mm}
\end{figure}

\textbf{Comparison with SOTA methods.} We first compare our method with competitive baselines on the validation set of Occ-ScanNet benchmark. As shown in Table~\ref{tab: quantitative comparison}, our vanilla DiScene already surpasses most existing methods. Solely through distillation and initialization strategies, it elevates the mIoU of our baseline model OPUS by $36.1\%$, from 28.70 to 39.06. This performance marginally exceeds that of OPUS$\dagger$, which requires a pre-trained depth model, while maintaining the highest inference speed at 23.2 FPS. This demonstrates the effectiveness of our approach in balancing accuracy and real-time efficiency. Furthermore, when integrating the depth branch, our model advances the mIoU of the baseline OPUS$\dagger$ from 38.96 to 47.17, establishing a new SOTA and outperforming the previous best method EmbodiedOcc by $3.7\%$. Crucially, it retains real-time capability at 10.5 FPS, further validating the superiority of our method.

To illustrate the performance of our method more intuitively, we also provide qualitative visualizations in Fig.~\ref{fig:vis}. Compared to the previous SOTA method EmbodiedOcc, our model demonstrates superior comprehension of geometry and semantics in complex and diverse indoor scenes. For instance, it accurately recognizes and reconstructs objects like the \textit{table} in the first column and the \textit{sofa} in the third column. Similarly, our approach outperforms the baseline method OPUS$\dagger$ in these scenarios, demonstrating enhanced capability in comprehending global scene structures and capturing finer local details. For example, our model successfully identifies the \textit{sofa} at a distance in the second column and the \textit{books} on the table in the last column, while existing methods fail in both cases. These findings underscore the efficacy of knowledge distillation in strengthening scene understanding capabilities.

\begin{table}[t]
\vspace{-1mm}
\centering
\caption{Effectiveness of knowledge distillation}
\vspace{-3mm}
\begin{tabular}[b]{l|cc|cc}
\toprule
Method & PDM & Param (M) & IoU & mIoU \\
\hline
Teacher & $\times$ & 379.1 & 52.79 & 48.42 \\
OPUS & $\times$ & 73.7 & 35.58 & 28.70 \\
DiScene & $\times$ & 73.7 & 43.68 & 39.06 \\
\hline
Teacher$\dagger$ & M3Dv2-G & 379.7 (1757.4) & 59.84 & 56.58 \\
OPUS$\dagger$ & DAv2 & 74.3 (172.0) & 45.62 & 38.96 \\
DiScene$\dagger$ & DAv2 & 74.3 (172.0) & 51.99 & 47.17 \\
\bottomrule
\multicolumn{5}{l}{We report learnable (w/o bracket) and total (w/ bracket) param.}\\
\multicolumn{5}{l}{M3Dv2 denotes Metric3D v2~\cite{hu2024metric3dv2}.}\\
\end{tabular}
\vspace{-4mm}
\label{tab:distillation}
\end{table}

\begin{table}[t]
\vspace{-2mm}
\centering
\caption{Ablation study of each component in DiScene}
\vspace{-3mm}
\begin{adjustbox}{width=0.88\linewidth}
\begin{tabular}[b]{cccc|cc}
\toprule
\multicolumn{3}{c}{MCKD} & \multirow{2}{*}{TGI} & \multirow{2}{*}{IoU} & \multirow{2}{*}{mIoU} \\
Query-Level & Prior-Level & Anchor-Level & & & \\
\hline
 & & & & 45.62 & 38.96 \\
 $\checkmark$ & & & & 48.32 & 42.83 \\
 & $\checkmark$ & & & 48.06 & 42.87 \\
 & & $\checkmark$ & & 48.19 & 42.82 \\
 & & & $\checkmark$ & 49.61 & 44.44 \\
 $\checkmark$ & & & $\checkmark$ & 50.16 & 45.27 \\
 $\checkmark$ & $\checkmark$ & & $\checkmark$ & 50.35 & 45.61 \\
 $\checkmark$ & $\checkmark$ & $\checkmark$ & $\checkmark$ & \textbf{51.99} & \textbf{47.17} \\
\bottomrule
\end{tabular}
\label{tab:ablation}
\end{adjustbox}
\vspace{-6mm}
\end{table}

\textbf{Effectiveness of knowledge distillation.} The results are illustrated in Table~\ref{tab:distillation}. Through knowledge distillation, we achieve substantial mIoU improvements of $36.10\%$ and $21.07\%$ for the student model under both settings. Concurrently, our approach reduces learnable parameters by over $80\%$ compared to the teacher model, with nearly $90\%$ total parameter reduction when incorporating the pre-trained depth model. These results validate the effectiveness of our method in balancing accuracy and computational costs, demonstrating strong suitability for practical deployment.

\begin{figure*}[t]
  \centering
  \includegraphics[width=0.9\linewidth]{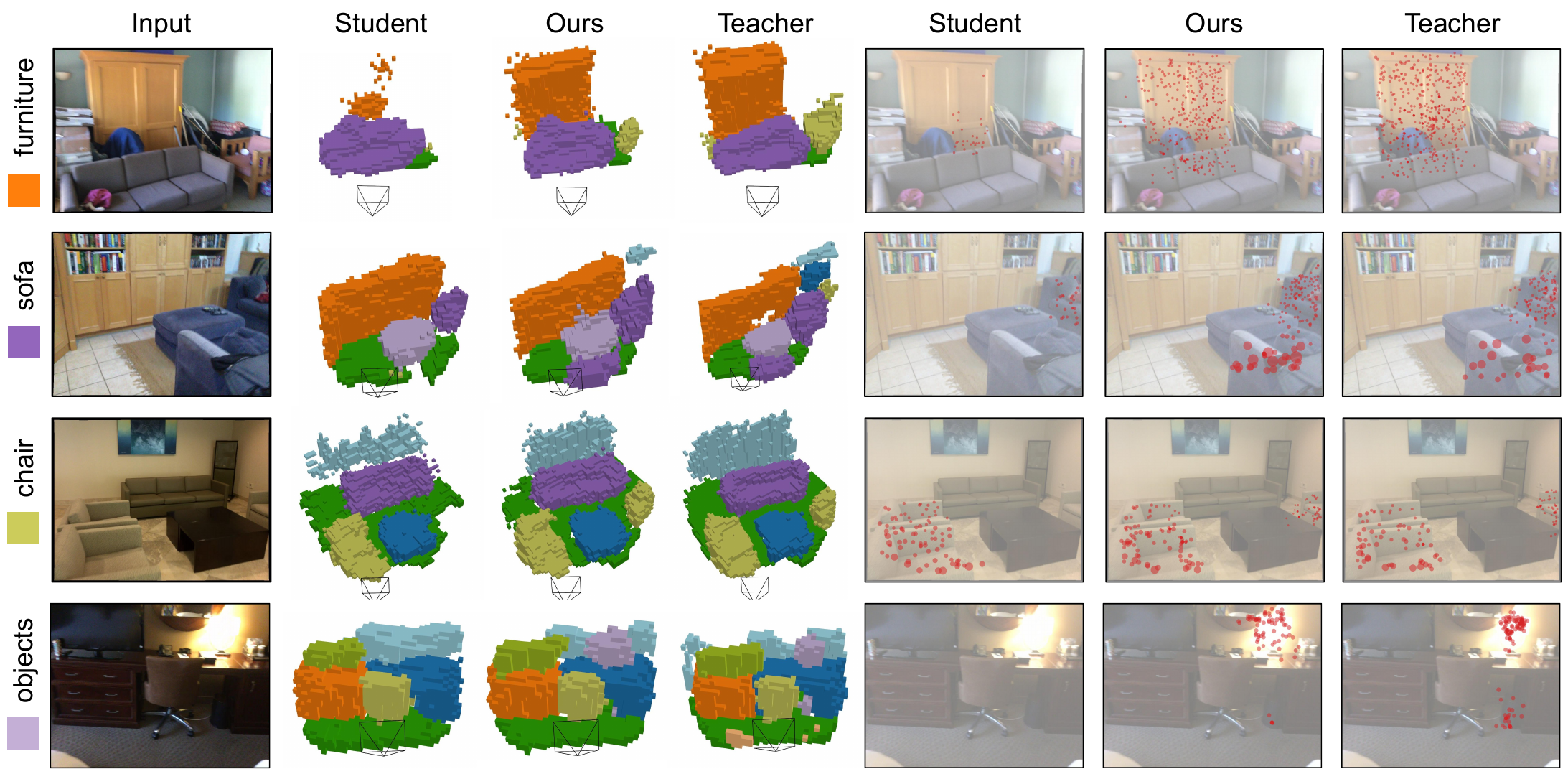}
  \vspace{-3mm}
  \caption{Visualization of occupancy predictions and activated query distributions across non-distilled student, distilled student, and teacher models. Activated queries are highlighted in red, with higher density near ground-truth regions indicating superior performance. The size of the circle represents the distance of the query center: larger circles are closer to the camera. We adjust the opacity of certain figures for better illustration. Best viewed in color.}
  \label{fig:distillation}
  \vspace{-7mm}
\end{figure*}

We further showcase the effectiveness of knowledge distillation in Fig.~\ref{fig:distillation}, which compares predictions from the non-distilled student, distilled student, and teacher models. For each row, we visualize occupancy predictions across models as well as the spatial distributions of activated queries for a specific semantic category. As demonstrated, the activated queries of the distilled student model exhibit significantly closer alignment with the teacher's spatial distribution, which is particularly evident in the first row. After distillation, we observe a substantial increase in the quantity of activated queries. These queries concentrate closer to ground-truth regions, accompanied by remarkably improved prediction accuracy compared to the non-distilled baseline. These results confirm that our distillation strategy enables the student model to effectively learn the teacher's spatial distributions and feature representations, thereby achieving performance gains. This validates both the correctness and efficacy of our Multi-level Consistent Knowledge Distillation framework.

\begin{table}[t]
\centering
\caption{Ablation study of distillation strategy}
\vspace{-3mm}
\begin{tabular}[b]{l|cc}
\toprule
Distillation & IoU & mIoU \\
\hline
$\times$ & 45.62 & 38.96 \\
\hline
Query-Level & \textbf{48.32} & \textbf{42.83} \\
\ \ \ \ w/ FLD & 47.07 & 41.70 \\
\ \ \ \ w/o EFA & 46.01 & 40.44 \\
\hline
Prior-Level & \textbf{48.06} & \textbf{42.87} \\
\ \ \ \ w/ FLD & 47.34 & 41.39 \\
\ \ \ \ w/o EFA & 46.45 & 40.40 \\
\hline
Anchor-Level & \textbf{48.19} & \textbf{42.82} \\
\ \ \ \ w/ FLD & 47.36 & 41.82 \\
\ \ \ \ w/o EFA & 45.73 & 38.45 \\
\bottomrule
\end{tabular}
\vspace{-6mm}
\label{tab:strategy}
\end{table}

\subsection{Ablation Study}

\textbf{Effects of each component.} To investigate the impact of each component in DiScene, we report the performance of each module in Table~\ref{tab:ablation}. When individually applying query-level, prior-level, and anchor-level distillation, we observe mIoU gains of 3.87, 3.91, and 3.86, respectively. This demonstrates the effectiveness of knowledge transfer at each distinct level. Furthermore, solely applying Teacher-Guided Initialization policy significantly boosts model performance by 5.48 mIoU, confirming its simplicity and efficacy. As all four components are progressively integrated into the framework, the model achieves steady performance improvements, culminating in a total gain of 8.21 mIoU. These findings underscore the importance and contribution of each component in our approach.

\textbf{Selection of distillation strategy.} In Table~\ref{tab:strategy}, we compare the impact of different distillation strategies on model performance. When replacing coarse feature-based distillation with fine-grained logit-based distillation (FLD), performance degradation is observed across all three levels. This decline is likely attributable to non-strict correspondence between point sets and voxels, where overly rigid fine-grained constraints may impede student learning. In contrast, coarse distillation imposes minimal restrictions on internal point distributions within the point set, thereby facilitating more effective model optimization. Furthermore, removing the encoder-level feature alignment loss causes significant performance drops at all levels, with the mIoU performance of anchor-level distillation even falling below that of the non-distilled baseline. These results validate that direct knowledge distillation between models with heterogeneous encoders suffers from substantial feature discrepancy, while our feature alignment loss effectively mitigates this issue.

\begin{table}[t]
\centering
\caption{Ablation study of different pre-trained depth model}
\vspace{-3mm}
\begin{tabular}[b]{l|c|cc}
\toprule
Model & FPS & IoU & mIoU \\
\hline
$\times$ & \textbf{23.2} & 35.58 & 28.70 \\
\hline
Depth Anything v1~\cite{yang2024depthanything} & 7.1 & 41.13 & 34.46 \\
Depth Anything v2~\cite{yang2024depthanythingv2} & \underline{10.5} & \underline{45.62} & \underline{38.96} \\
\hline
Metric3D v2-S~\cite{hu2024metric3dv2} & 8.4 & 43.77 & 37.82 \\
Metric3D v2-G~\cite{hu2024metric3dv2} & 1.2 & \textbf{47.45} & \textbf{41.17} \\
\bottomrule
\end{tabular}
\label{tab:PDM}
\vspace{-6mm}
\end{table}

\textbf{Selection of pre-trained depth model.} Table~\ref{tab:PDM} presents model performance using different pre-trained depth models. We evaluated two models producing relative depth estimations, Depth Anything v1~\cite{yang2024depthanything} and v2~\cite{yang2024depthanythingv2} (both \textit{fine-tuned} on indoor scenes to get metric outputs), alongside Metric3D v2~\cite{hu2024metric3dv2}, a \textit{zero-shot} model producing metric estimations. These results reveal that models integrating Depth Anything v2 achieve the fastest inference speed among all depth-enhanced variants while delivering the second-best mIoU performance. Conversely, models utilizing Metric3D v2-G attain peak accuracy but suffer from severely constrained inference speeds. Based on these observations, our DiScene$\dagger$ strategically employs Depth Anything v2 in the student model to strike an optimal accuracy-speed balance, while adopting Metric3D v2-G in the teacher model to ensure demonstrably more robust performance.

\begin{figure*}
  \centering
  \includegraphics[width=0.9\linewidth]{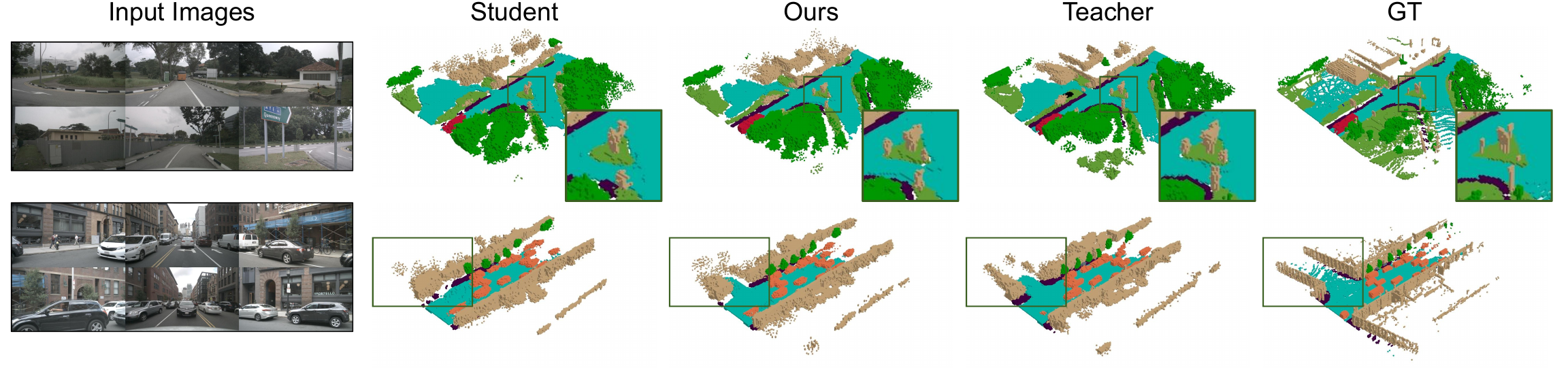}
  \vspace{-5mm}
  \caption{Qualitative results of occupancy prediction on the Occ3D-nuScenes validation set. The boxes highlight finer local detail capture (Row 1) and enhanced scene structure reconstruction (Row 2) achieved by our distilled model.}
  \label{fig:robust}
  \vspace{-7mm}
\end{figure*}

\subsection{Robustness Analysis}

In this section, we investigate the robustness of our distillation strategy in outdoor driving scenarios.

\textbf{Experimental setup.} Our experiments are conducted on the Occ3D-nuScenes benchmark~\cite{tian2023occ3d}, which provides dense semantic occupancy annotations for the widely used nuScenes dataset~\cite{caesar2020nuscenes}. 
Each voxel is annotated with 18 classes (17 semantic classes and 1 free). Following~\cite{wang2024opus}, we use mIoU and RayIoU as evaluation metrics. Implementation details remain consistent with Section~\ref{sec: implement}.

\textbf{Quantitative results.} As evidenced in Table~\ref{tab:occ3d}, the integration of distillation and initialization strategies yields a $6.92\%$ mIoU improvement and $5.79\%$ RayIoU gain over the baseline, while reducing learnable parameters by nearly $80\%$ compared to the teacher model. These results demonstrate the efficacy of our method in outdoor scenarios, achieving performance gains with reduced computational overhead, thus confirming its robustness in both indoor and outdoor environments.

\begin{table}[t]
\centering
\caption{Performance on the Occ3D-nuScenes dataset}
\vspace{-3mm}
\begin{tabular}[b]{l|c|cc|ccc}
\toprule
Method & Param (M) & mIoU & $\text{RayIoU}$ & \multicolumn{3}{c}{$\text{RayIoU}_{\text{1m, 2m, 4m}}$} \\
\hline
Teacher & 382.9 & 33.82 & 37.9 & 30.9 & 38.9 & 43.9 \\
OPUS & 77.5 & 28.31 & 32.8 & 26.2 & 33.7 & 39.0  \\
DiScene & 77.5 & 30.27 & 34.7 & 27.9 & 35.5 & 40.7 \\
\bottomrule
\end{tabular}
\vspace{-5mm}
\label{tab:occ3d}
\end{table}

\begin{figure}[ht]
  \centering
  \includegraphics[width=0.9\linewidth]{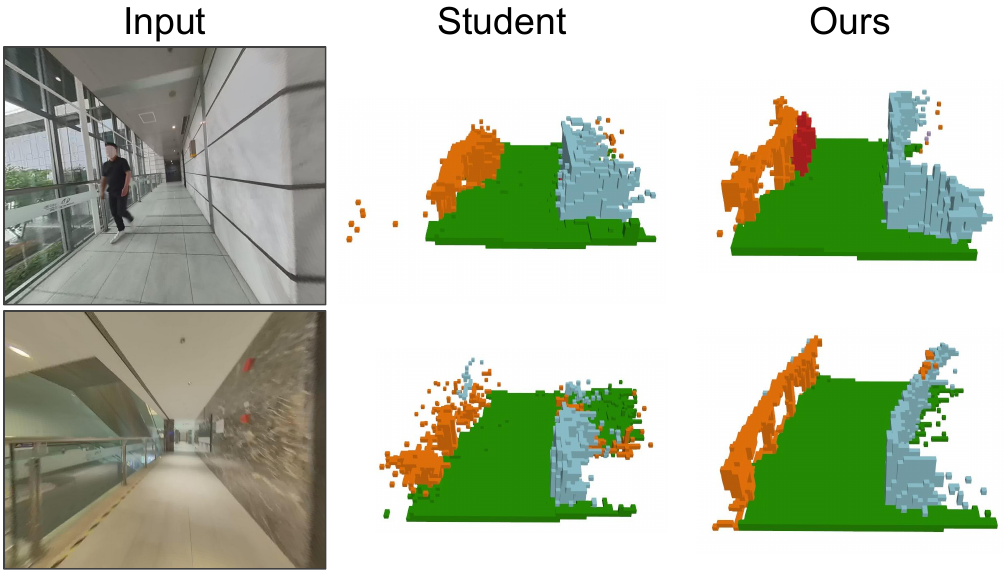}
  \vspace{-3mm}
  \caption{Qualitative results of occupancy prediction on self-collected in-the-wild datasets. Our method demonstrates enhanced capabilities in geometric and semantic understanding.}
  \label{fig:in-the-wild}
  \vspace{-3mm}
\end{figure}

\textbf{Qualitative results.} We further visualize the prediction results in Fig.~\ref{fig:robust}. Our distilled model demonstrates markedly superior capabilities over the baseline in capturing local details and obtaining holistic structures. In the first row, the baseline erroneously predicts the distribution of \textit{poles} at the intersection center, while the second row reveals its inaccurate \textit{road structure} reconstruction. These limitations are effectively resolved through distillation, yielding predictions that closely align with the teacher model and exhibit enhanced scene comprehension capabilities. Collectively, these results validate the effectiveness and robustness of our method across diverse perception scenarios. Furthermore, the demonstration of in-the-wild scenes in Fig.~\ref{fig:in-the-wild} indicates the versatility of our approach.

\subsection{Failure Cases}
\begin{figure}[ht]
  \centering
  \includegraphics[width=0.88\linewidth]{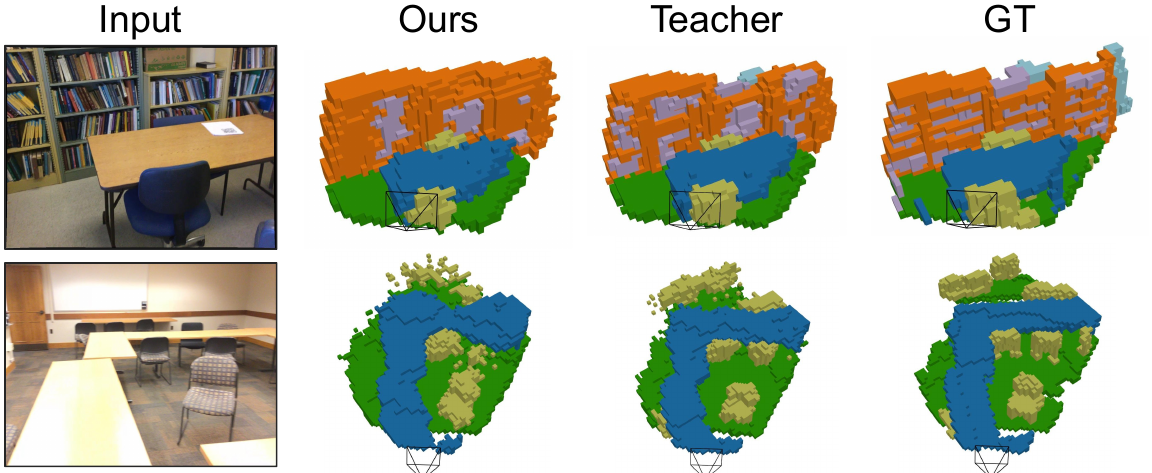}
  \vspace{-3mm}
  \caption{Failure cases of our approach.}
  \label{fig: case}
  \vspace{-6mm}
\end{figure}

Fig.~\ref{fig: case} illustrates several failure cases of our approach, in which the student model still struggles to effectively learn from the teacher through distillation. These cases typically occur when objects of a certain category exhibit both high density and large spatial distribution in the image, often accompanied by partial occlusion, such as the \textit{books} on the bookshelf in the first row and the \textit{chairs} in the second row. Such scenarios provide an abundance of intricate visual cues, which significantly increases the difficulty of learning both spatial distributions and feature representations, thereby limiting the efficacy of knowledge distillation. We believe that this issue could be addressed by incorporating instance-level priors, meriting deeper investigation in future work.

\section{Conclusion} 
\label{sec:conclusion}

In this paper, we present DiScene, a novel framework for sparse
query-based occupancy prediction. We propose Multi-level Consistent Knowledge Distillation, a hierarchical distillation strategy incorporating coordinated distillation across multiple complementary levels. This approach ensures consistent feature alignment and robust knowledge transfer, significantly boosting the performance of student model. Moreover, we introduce a Teacher-Guided Initialization policy that significantly accelerates convergence and enhances model performance without incurring additional computational costs. Our method optimally balances real-time efficiency with prediction accuracy, establishing new SOTA performance on the Occ-ScanNet benchmark while demonstrating robustness across diverse environments. We hope that DiScene can establish a practical paradigm for enhancing 3D perception in resource-constrained and complex indoor scenarios.

\bibliographystyle{IEEEtran}
\bibliography{IEEEabrv,reference}

\end{document}